# RGB cameras failures and their effects in autonomous driving applications

Andrea Ceccarelli, Francesco Secci

**Abstract**— RGB cameras are one of the most relevant sensors for autonomous driving applications. It is undeniable that failures of vehicle cameras may compromise the autonomous driving task, possibly leading to unsafe behaviors when images that are subsequently processed by the driving system are altered. To support the definition of safe and robust vehicle architectures and intelligent systems, in this paper we define the failure modes of a vehicle camera, together with an analysis of effects and known mitigations. Further, we build a software library for the generation of the corresponding failed images, and we feed them to six object detectors for mono and stereo cameras and to the self-driving agent of an autonomous driving simulator. The resulting misbehaviors with respect to operating with clean images allow a better understanding of failures effects and the related safety risks in image-based applications.

**Index Terms**— autonomous driving; RGB camera; failures; obstacles detection; driving simulation; image-based applications

——————————— ◆ ———————————

## 1 INTRODUCTION

Autonomous driving is attracting growing attention in recent years, with ever-increasing demand and investments from the industry [17]. The objective of an autonomous driving system is to drive by itself without requiring help from a human: the vehicle detects the environment, locates its position, and operates to safely reach a specified destination.

Sensor technology, data-fusion, and inference algorithms as Artificial Intelligence and Machine Learning (AI/ML) applications are the enabling technologies that play a cornerstone role in autonomous driving systems. These are involved in many of the essential tasks for safe driving such as sensor-fusion, environment representation, scene understanding, semantic segmentation, tracking, object detection, and recognition [15].

Amongst sensors, the RGB (red, green, and blue) camera is acknowledged as the most commonly used and an irreplaceable one [15]. In fact, despite cameras have the known disadvantages of strong sensitivity to external illumination and limited field of view, visual recognition systems are amongst the most solid applications of autonomous driving [33]. Vehicle cameras are already exploited in many applications such as traffic sign recognition, lane detection, obstacle detection, etc. [15], [16], [17]. Additional prospective applications are being researched; for example, at intersections, knowing the location of pedestrians and bicyclists can allow the car to make sophisticated precedence decisions [16]. Also, cameras are amongst the cheapest solutions to build autonomous driving systems that can sense the surroundings [17].

When the images provided by the camera are degraded, fatal accidents may occur. The trained agents of the AI/ML applications responsible for the elaboration of inputs may rely on biased data and consequently lead to wrong (unsafe) decisions.

Several works explore how to make the trained agents robust to artificially crafted or accidentally manipulated input images [35], [36], [37], and how to secure a camera from direct attacks that may disrupt the proper behavior of the camera itself [18], [19]. In fact, even slight alterations of the images may alter the output of the trained agent [39]. However, few or no works focused on the attentive identification of a realistic and complete set of accidental modifications of images from a failed RGB camera. Rather, we believe such analysis would contribute to a better understanding of the effects of cameras failures and the related risks, for example when cameras are applied in safety-critical domains such as autonomous driving. Consequently, in this paper we propose a systematic analysis of the possible malfunctions of a camera and a detailed description of the related failures modes. The systematic analysis can bring benefits to system design, during which hazards are investigated and countermeasures selected. In addition, we provide software libraries for the reproduction of the failures, and we experiment with them, such that this study can support the robustness assessment of image-based AI/ML applications.

The contribution of this paper is organized in four parts. *First*, we identify failure modes of vehicle cameras in the domain of autonomous driving, by analyzing the different failures, their causes, and their potential effects on the system. We achieve this goal by applying an FMEA (Failure Modes and Effects Analysis, [14]) on the components of an RGB camera, assuming the camera is located on a vehicle. *Second*, we review existing mitigations; this review shows that while mitigations exist for essentially any individual failure, an orchestrated approach to the whole set of failures is still missing. *Third*, we exercise trained agents for object detection on failed images, to measure the performance drop due to the occurrence of failures. To realize this, we alter images from the KITTI Vision Benchmark Suite for

————————————————

• *Andrea Ceccarelli is with the Department of Mathematics and Informatics, University of Florence, Viale Morgagni 65, 50134, Florence, Italy. E-mail: andrea.ceccarelli@ unifi.it.*
• *Francesco Secci is with the Department of Mathematics and Informatics, University of Florence, Viale Morgagni 65, 50134, Florence, Italy.*

autonomous driving [75], [76] according to our failure modes, using a python library that we developed and that is publicly available at [9]. At this point, we execute six object detectors on the modified images (that we make available at [10]), measuring the resulting detection capability in terms of Average Precision [77]. Last, to bring further evidence on the failure criticalities in the autonomous driving domain, we inject failures in a self-driving agent for autonomous driving in a simulated environment: this shows the effects of camera failures in applications that repeatedly acquire frames and take decisions.

This work is an extended version of [78]: we detail the five most noticeable additional contributions with respect to [78], in what we believe is increasing order of relevance. To facilitate the reader, we also explicitly mention the related Sections of this work. First, we present a revised description of the failures, including also a notation that facilitates the understanding of each failure (Section 3 and Table I). Second, we present a revised version of the python library that allows simulating the camera failures. In addition to technical improvements to generate failed images, the library allows simulating a new failure (flare), and includes the possibility to exercise multiple configurations; for example, in total, we experiment with 130 configurations (Section 4 and Table III; the source code is available at [9], and the produced image dataset is available at [10]). Fourth, we describe the methodology applied in our experimental campaign (Section 4 and especially Section 4.4), which acts as reference to build analogous campaigns that evaluate the robustness of trained agents. Fifth and most important, Section 4 studies the effects of failures on six trained agents (six object detectors). This is a relevant addition with respect to [78], which just showed that camera failures may cause the misbehavior of a single self-driving agent. In this paper, we show common trends of the six trained agents when affected by the different failures, and we measure the impact of the failures in terms of reduction of detection performance. Thanks to this, we can provide quantitative indications on the most dangerous failures (and on those that resulted almost negligible). Summarizing, in addition to [78], this paper provides i) substantial evidence that camera failures should be considered a relevant threat; ii) reference data on the impact of the different failures; iii) a methodology to evaluate robustness of trained agents and in particular object detectors.

We observe that Section 5 reports a summarized view of the experimental campaign in [78]. Section 5 describes the injection of camera failures in a simulated self-driving agent under different operating conditions, to show the effects of camera failures in applications that repeatedly acquire frames and take decisions. In [78], the reader may find results of additional runs, here not reported. At the light of the novel results introduced in this paper and discussed above, which explain the risks connected to camera failures, we believe the concise description reported in Section 5 is fully appropriate to provide the reader with all the relevant insights on such experimental campaign.

The rest of the paper is organized as follows. In Section 2 we present the fundamentals that are at the basis of our work. In Section 3 we detail the identified failures, their effects on the output image, and possible mitigations. In Section 4 we execute object detectors on altered images from the KITTI dataset. In Section 5 we inject the failures in the frontal camera of a simulated vehicle that is driving autonomously. Finally, in Section 6 we review related works and in Section 7 we define conclusions.

## 2 BACKGROUND NOTIONS

We present background notions that are at the basis of our work. In Section 2.1 we define the architectural structure of an RGB camera (simply called camera, from now on) that we use as reference in our work, and in Section 2.2 we describe the FMEA methodology that we apply to identify camera failures.

### 2.1 Architecture of a Camera

We consider a camera structured in five components (Fig. 1): lens, camera body, Bayer filter, image sensor, and ISP (Image Signal Processor) [27]. These five components contribute to the creation of the output image.

**Lens.** Photographic lenses are devices capable of collecting and reproducing an image [7]. The lens is the component that has the greatest impact on the quality of the images. The photographic lens can be composed of one or more lenses and/or reflectors as for example systems of concave and convex mirrors, often also combined with diopters. The fundamental factor that distinguishes one lens from another is primarily the focal length. A second factor that characterizes the lens is brightness. A third distinguishing factor is macros: a macro lens can focus from infinity to 1:1 magnification that is, the size of the image in real life is the same as it is reproduced on the sensor. Another relevant factor is the focus: this can be manual or automatic. The lens also contains a minimum of electronics, necessary for the focus motor (when automatic) and for zooming [20].

**Camera Body.** The camera body is the container of all the electronics of the camera. The Bayer filter, the image filter, and the ISP are here contained. Typically, the functions of the camera body are securing the device and protecting inner components from exposure and contact with the outside. For example, the case protects the sensor from light and other possible sources of damage.

**Bayer Filter.** The Bayer filter (or Bayer pattern) is used for the acquisition of digital images [71]. The photodiodes in an image sensor are color-blind by nature: they can only register shades of gray. To obtain the color in the image, they are covered with different color filters: red, green and

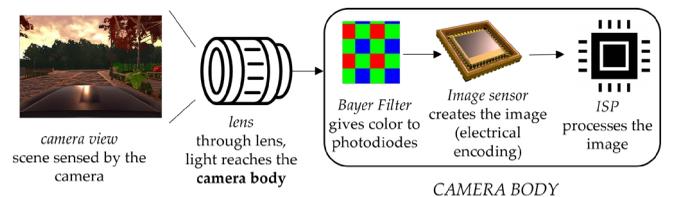

*camera view*
scene sensed by the camera

*lens*
through lens,
light reaches the
**camera body**

*Bayer Filter*
gives color to
photodiodes

*Image sensor*
creates the image
(electrical
encoding)

*ISP*
processes the
image

*CAMERA BODY*

Fig. 1. A camera and its components: the Lens, and the Camera Body composed of Bayer Filter, Image Sensor, and Image Signal Processor.

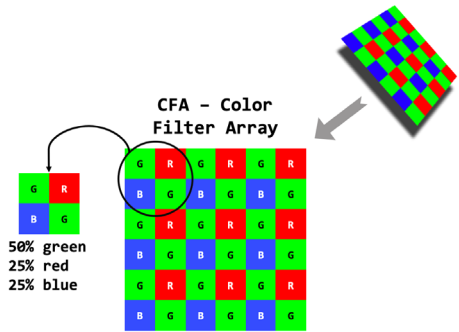

Fig. 2. Scheme of a Bayer filter. On the left, a sample cell of 2x2 photosites (BGGR pattern). In digital imaging, a color filter array (CFA), or color filter mosaic (CFM), is a mosaic of tiny color filters placed over the pixel sensors of an image sensor to capture color information [41].

blue (RGB) according to the model designated by the Bayer filter. This filter groups the sensors for the three fundamental colors RGB in cells of 2x2 photosites: each cell contains two green elements, one red element and one blue element (Fig. 2). Each pixel is filtered to register only one of the three colors: to obtain a color image, a demosaicing algorithm interpolates a set of complete red, green and blue values for each pixel. This algorithm uses the surrounding pixels to estimate the value of a particular pixel [70].

**Image Sensor.** The image sensor is the transducer that converts the image into its representation or electrical coding. Essentially, it is a silicon chip capable of capturing and measuring light i.e., the number of photons which reach the chip. The sensor surface is made up of millions of tiny receivers arranged in a regular grid; these receivers, also called photosites, are the microsensors that carry out the conversion from photons to electrons. Each receiver can supply an electrical charge proportional to the number of photons that hit it. The detected charge is then converted by a special analog-digital conversion circuit into a numerical value. Each of the values obtained from the photosites will constitute a pixel of the obtained image.

There are currently two types of image sensors on the market: CCD (Charge Coupled Device) and CMOS (Complementary Metal Oxide Semi-conductor). Both are based on the concept of converting the charge of each photosite into a digital format using an ADC (Analog to Digital Converter), but they differ in how the information is processed. In fact, for CCD sensors, the information on the charge is taken from the photosites row by row and stored in a register whose content is passed to an amplifier and subsequently to the ADC. After the row has been fully processed, it is eliminated from the exit register and the next row that undergoes the same treatment is loaded. Instead in CMOS sensors, together with photosites, transistors have been integrated which perform amplification and conversion of the charge into voltage. Using a matrix structure, it is possible to individually select each photosite through its [*row, column*] coordinates, and then send the voltage to an ADC that performs the conclusive digital conversion [6], [42]. Typically, the image sensor outputs a *raw* file, which is further processed by the ISP [93].

**ISP.** The Image Signal Processor is a type of specialized media processor or Digital Signal Processor (DSP) used for image processing in digital cameras [42]. The ISP produces the digital image (simply called image when clear from context); its functions are multiple, such as: demosaicing, correction of the image sensor, noise reduction, image sharpness correction, resizing the image, lens distortion correction, chromatic aberration correction, image compression and JPEG encoding, video compression, and more [42].

## 2.2 Principles of FMEA

The Failure Mode and Effects Analysis (FMEA, [14]) is a widely-used reliability management technique designed to identify potential failures of a component or a process, understand the effects of these failures, assess the associated risk, and ultimately classify problems in terms of importance [4]. This allows choosing and implementing corrective actions to address the most serious potential failures. Usually performed analytically, an FMEA is composed of four stages [4], [5]: i) identify all known or potential failure modes of a system; ii) confirm the *causes* and *effects* of each failure; iii) rank the recognized failures by their *risk*, defined as a combination between the probability of occurrence of a failure and the severity of the latter; iv) take remedial actions for the highest risky failures.

The assumption underlying the application of the FMEA is the principle according to which the risk is related not only to the probability that a failure occurs, but also the seriousness of its consequences and ability to avoid or mitigate it. Ultimately, FMEA provides a knowledge base on the possible failures and their effects, which can be used for future troubleshooting activities [4], [5], [14]. This matches the objectives of our work and it is the reason why this technique was selected.

## 3 ANALYSIS OF CAMERA FAILURES

We consider a frontal camera of a vehicle organized in the five components discussed in Section 2.1, and we assume that output images are then processed by image-based AI/ML applications. We exercise the FMEA on the components of the camera. The application of the FMEA identifies the failure modes of the camera components, the cause of such failures, and their effect at the camera level i.e., on the output image. Further, we complement this list with a literature review on camera failures, to assure that no relevant failure modes are left out. With respect to the usual analytical application of FMEA, due to the absence of reference data, we could not associate credible ratings on the risk matched to an individual failure (this is in line with acknowledged limitations of risk ratings in FMEA [4]). However, we mitigate this gap through Section 4 and Section 5, via the execution of image-based applications.

In Section 3.1 we list the failures in alphabetic order; we assign an evocative name to each failure, that we will use in the rest of the paper. In Section 3.2 we summarize the failures, and we report state-of-the-art mitigations. To facilitate the reader, we associate an acronym to each failure, and we describe it with a synthetic identification of the

TABLE I

FAILURES ACRONYM AND SYNTHETIC DESCRIPTION.

| Failure | NAME(Component, Input, Output): Effects |
|---------|------------------------------------------|
| Banding | BAND(image sensor, light, raw): altered image |
| Brightness | BRIGHT(lens, light, light): altered image |
| Blur | BLUR(lens, light, light): altered image |
| Brackish/Salt-Water | BRACK(lens, light, light): altered image<br>BRACK(camera body, light, none): no image |
| Bright Lines | BRLINES(image sensor, light, raw): altered image |
| Broken Lens | BRLE(lens, light, light): altered image |
| Broken VR | BRVR(lens, light, light): altered image |
| Condensation | COND(lens, light, light): altered image<br>COND(camera body, raw, none): no image |
| Dead Pixel | DEAPIX(image sensor, light, raw):  altered image |
| Dirty | DIRTY(lens, light, light): altered image |
| Electrical Over-load | ELOV(camera body, light, image): altered image<br>ELOV(camera body, light, none): no image |
| Flare | FLARE(lens, light, light): altered image |
| Heat | HEAT(lens, light, light): altered image<br>HEAT(camera body, light, image): altered image<br>HEAT(camera body, light, none): no image |
| Ice | ICE(lens, light, light): altered image<br>ICE(camera body, light, none): no image |
| No Action | NOACT(ISP, raw, none): no image |
| No Bayer Filter | NBAYF(Bayer filter, light, raw): altered image |
| No Chro. Aber. | CHROMAB(ISP, raw, image): altered image |
| No Demosaicing | DEMOS(ISP, raw, image): altered image |
| No Lens Dist. | DISTOR(ISP, raw, image): altered image |
| No Noise Red. | NOISE(ISP, raw, image): altered image |
| No Sharpness | SHARP(ISP, raw, image): altered image |
| Rain | RAIN(lens, light, light): altered image |
| Sand | SAND(lens, light, light): altered image<br>SAND(camera body, light, image): altered image |
| Spots | SPOTS(image sensor, light, raw): altered image |
| Water | WATER(lens, light, light): no image<br>WATER(camera body, light, none): no image |
| Wind | WIND(lens, light, light): altered image<br>WIND(camera body, light, image): altered image |

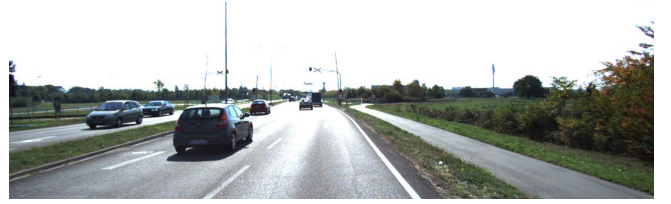

Fig. 3. Sample image from the KITTI Vision Benchmark [75], [76] suite, used as references to explain the failures in Fig. 4.

4a shows the Banding failure injected into Fig. 3. Only a portion of the banded image is shown, to make the banding effects visible to the naked eye. Banding failures manifest in the image sensor.

**Brightness.** These failures represent the brightness alteration, from its minimum (black image) to its maximum (white image) limits, that can happen with the breakdown of fundamental components of the lens such as the shutter, the diaphragm, or the iris  (Fig. 4b). For example, if the shutter has a malfunction that does not allow the entrance of the correct amount of light through the lens, brightness could be altered, from entirely black to entirely white. These brightness failures manifest in the lens component.

**Blurred**. Blur may occur if the image captured by the device is not in focus (Fig. 4c) [69]. Especially in autonomous driving, it is of fundamental importance that the captured images are of good quality and therefore in focus. This is clear if we think that the image-based AI/ML applications make decisions regarding the vehicle's movement, based on the information content of the camera images. Blurred failures manifest in the lens component.

**Brackish/Salt-Water.** The brackish phenomenon is common in coastal areas and puts a strain on the durability of the materials if not treated with suitable products and/or not maintained over time [6]. The corrosive power of water and air given by the substantial percentage of salt may damage the lens and the camera body itself, to the extent that external agents may enter the circuits [43]. In the worst scenario, this can affect image acquisition. The image may be altered in various ways; we refer to the description of all the other failures effects (Fig. 4) for the complete characterization of the Brackish/Salt-Water effect. This failure manifests in the lens and camera body.

**Bright Lines.** This failure is very rare with the current knowledge and technology. The produced images could show bright vertical and/or horizontal lines, also clearly distinguishable with the naked eye. The cause of these lines is due to the use of LIDARs: this laser technology emits a light intensity (not visible to the human eye) that can seriously damage the camera's image sensor. This failure may manifest in the image sensor.

**Broken Lens.** One or more internal or external lenses may break, for example, because of mechanical stresses due to vehicle jolts or the impact with gravel throw-up by the tires of nearby vehicles.   The camera regularly outputs the target component, its relevant input, and the output. In Table I, we summarize each failure using the notation:

$$NAME(component, input, output): effect$$

where i) *NAME* is the failure acronym, ii) *component* is either *lens*, *camera body*, *Bayer filter*, *image sensor*, or *ISP*, iii) *inputs* and *outputs* are respectively inputs and outputs of the component when the considered failure is in place, and they can be *none* (in case there is no output), *light*, *raw*, and *image*, and iv) *effect* describes the failure effect at the camera level i.e., the camera output that may be used by image-based applications, and it can be *no image* or *altered image*. Further, to give a visual understanding of the effects of the failures, we apply them on a reference figure (Fig. 4).

### 3.1 List of identified failures

**Banding.** In this failure, many parallel horizontal and/or vertical lines become visible in the produced image. The lines are more visible when looking at the darker colors, although they can be also perceived on lighter ones. Figure

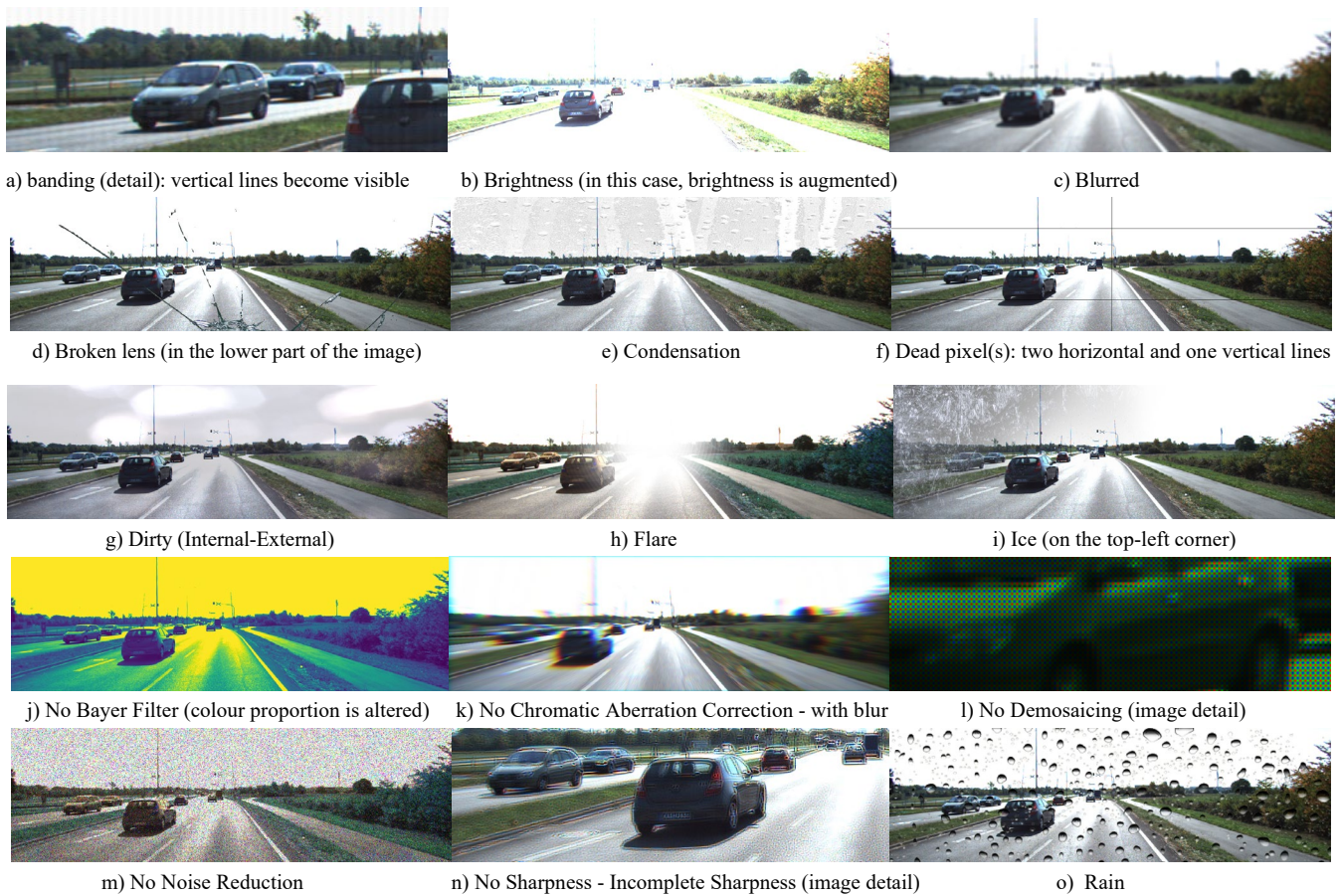

a) banding (detail): vertical lines become visible

b) Brightness (in this case, brightness is augmented)

c) Blurred

d) Broken lens (in the lower part of the image)

e) Condensation

f) Dead pixel(s): two horizontal and one vertical lines

g) Dirty (Internal-External)

h) Flare

i) Ice (on the top-left corner)

j) No Bayer Filter (colour proportion is altered)

k) No Chromatic Aberration Correction - with blur

l) No Demosaicing (image detail)

m) No Noise Reduction

n) No Sharpness - Incomplete Sharpness (image detail)

o) Rain

Fig. 4. Visual effect of different failures on the output image of a frontal camera. Failures are applied to the image in Figure 3.

image, but it will include an additional line (in case of a scratch) or more complicated patterns (Fig. 4d). This failure manifests in the lens component.

**Broken VR.** In this case, the malfunction of the component that deals with the reduction of vibrations (VR) is considered. This is located in the lens and it is common on many camera models. Its malfunction causes out-of-focus images: for this reason, the effects of this failure are similar to the Blurred ones (Fig. 4c). Broken VR manifests in the lens component.

**Condensation.** When the outside air temperature drops sharply, condensation may appear on the lenses. Condensation, or humidity, degrades the images (Fig. 4e). The image is acquired, but it may have defects due to halos on the lenses [72]. If humidity penetrates inside the camera body, it may cause malfunctions and it may also preclude the entire operability of the device. Condensation failure manifests in the lens and the camera body.

**Dead Pixel.** In this failure, the output images have one or more defects of pixel size. While this failure manifests in the image sensor, we call it dead pixel, because it has the same visual effect of the common failure that can be noticed on LCDs (Liquid Crystal Displays), when a pixel stops working properly and it appears as a black spot on the screen. A single dead pixel may not preclude the good interpretation of the captured images by the AI/ML applications, despite a deliberately modified pixel may do so [39]. Obviously, several dead pixels (e.g., an extreme case is in Fig. 4f) have higher chances to drop the accuracy of the AI/ML application that uses the image.

**Dirty (Internal – External).** This failure (Fig. 4g) concerns debris of various kinds and sizes (most typically, dust and dirt) which deposits on the internal or external lenses [72]. The most significant difference is that the external dirt can be removed, usually by cleaning the first lens of the objective (i.e., the most external lens). Instead, removing the internal dirt requires more time and, sometimes, specialized personnel [44]. Dirty-related failures manifest in the lens component.

**Electrical Overload.** The excessive and dangerous temperature increase of the conductors due to an electrical overload could damage, and most likely break, the electronical parts in the camera body. A device with this generic electrical problem may stop working or find itself in a state where images cannot be processed [44]. The resulting effect is that the images are produced incorrectly or, most likely, not produced at all. The image may be altered in a multitude of ways; we refer to the description of all the other failures for the complete identification of possible effects of electrical overload. Electrical overload may manifest in the camera body.

**Flare.** The structure itself of the lenses group creates flare (Fig. 4h). Flare is due to the reflection of the sun or other light sources on the lenses. The resulting image shows one or more spots of various colors, placed on an imaginary line. Obviously, these spots can cover details in the images: the subsequent processing steps, even if the image is captured correctly, will be influenced by the presence of these spots. Flare is difficult to eliminate also with modern technology, which sets a series of lenses slightly spaced one from each other, and each with a specific task. Flare failure manifests in the lens component.

**Heat.** This type of failure relates to the heat that the lens or camera body can suffer in their operational life. In extreme cases, excessive heat could lead to the evaporation of the lubricating liquids of the moving parts (e.g., zoom). As a result, the use of the zoom (if present) and of other subcomponents that are intended to make the image as clear as possible (e.g., the focus tools) may be precluded. As for Brackish/Salt-Water, the image may be altered in a multitude of ways; we refer to the description of all the other failures for the complete identification of possible effects. The heat failure manifests in the lens and camera body.

**Ice.** Ice can be the cause of several camera malfunctions. It can break the external materials of the camera lens and camera body. Furthermore, the external lens can be covered with a blanket of ice that prevents the acquisition of images (Fig. 4i). The Ice failure manifests in the lens and the camera body.

**No Action.** A failed ISP does not respond and therefore the processing of the acquired image does not take place: the image remains in raw format, without any type of processing. The camera does not transmit anything to the other processing components of the vehicle. This failure manifests in the ISP.

**No Bayer Filter.** Without a Bayer Filter properly functioning, the produced image would result in wrong colors (for example, in Fig. 4j, colors are altered using the RGB coefficients from ITU-R BT.601 [79] to generate achromatic images). The following phases would unavoidably process chromatically wrong images. This failure manifests in the Bayer filter.

**No Chromatic Aberration Correction - Incomplete Chromatic Aberration Correction.** It refers to the case that the ISP fails to (fully or partially) apply the removal of chromatic aberration on the acquired image. In optics, axial chromatic aberration is a defect in the formation of the image, due to the different refractive values of the light wavelengths that pass through the optical instrument [67]. It is a defect that may affect all optical lens systems, to varying degrees. This failure leads to images with colored halos on the edges of the subjects: the images show "fringes" of various colors (mostly purple) and a sort of blur (Fig. 4k). This failure manifests in the ISP.

**No Demosaicing - Incomplete Demosaicing.** In No Demosaicing, we consider the case in which the image is acquired in RAW format (Fig. 4l). This means that the demosaicing process has not been carried out and therefore the image is presented with each pixel containing a red, green, or blue value. In this case, the Bayer array has not yet been interpreted and the image is more pixelated than normal. The prevalence of the greenish hue, also visible in Fig. 4f, is due to the high percentage of green parts, which is double the percentage of the red and blue parts, following the BGGR pattern in the Bayer filter as previously shown in Fig. 2. This failure manifests in the ISP.

**No Lens Distortion Correction - Incomplete Lens Distortion Correction.** This failure may affect only vehicles which mount cameras with wide-angle lenses [21], [73], which tend to deform the image. In this case, the captured image appears as mapped around a sphere that is more protruding towards the observer, in the center of the image. If the conversion to normal proportions (natural symmetric) is not successful, the image may freeze in this processing phase and the system may be in a stalled state. Alternatively, if the output image is incorrectly processed, the AI/ML applications may use images with surrounding objects of distorted proportions and shapes. This failure manifests in the ISP.

**No Noise Reduction - Incomplete Noise Reduction.** The device captures the image, but during the processing phases (noise reduction) there is an error that prevents the correct removal of the noise e.g., Fig. 4m. This failure manifests in the ISP.

**No Sharpness - Incomplete Sharpness.** In this case, the processing of the captured images fails during the sharpness correction phase. This affects the ability of a camera to identify and define the separation limit between two contiguous areas that have different brightness and/or color (Fig. 4n). This failure manifests in the ISP.

**Rain.** It refers to the case in which there are small spots on the images due to the deposit of water drops on the external lens [72] (Fig. 4o). The elimination of these stains can be considered trivial when the vehicle is parked and without rainfall. However, given the weather variability that a vehicle can encounter, we cannot only consider such simple case. This failure manifests in the lens.

**Sand.** Because of sand, there may be possible corrosion of the external sub-components of the lens (percentage of salt in the sand), with the consequent introduction of external agents inside the device, and to the extent that the camera may not capture exact images. Sand could block sub-components that have the purpose of making the image as clear as possible (tools for focusing, zooming, etc.). The effect on the output image (when sand is on the lens) is similar to Dirty Internal-Dirty External (Fig. 4g). This failure manifests in the lens and camera body.

**Spots.** This failure occurs when small particles of dust (or another type of material) settle on the Image Sensor. This deposit means that small spots, or shadows, are visible above the light colors on the output image. Such shadows are mostly circular and are very common for amateur photography, particularly when using multiple lenses. In fact, while operating on the camera for maintenance, external agents of imperceptible size may enter and settle on the exposed Image Sensor [74]. The failure effect is similar to the Dirty Internal-Dirty External (Fig. 4g). This failure manifests in the image sensor.

**Water.** If water enters the lens or the camera body, the electrical components can fail and most likely no longer acquire images, or acquire them without any content [72]. This failure manifests in the lens and camera body.

**Wind.** We consider those parts of the component with cavities that, due to the force of the wind (while the vehicle is in motion or parked), could lead to minimal external damage and the subsequent infiltration of various agents inside the camera. The acquisition of images could therefore be incorrect: lens may move and images may be shifted or cut, etc. This can be considered very rare for a vehicle camera. This failure manifests in the lens and camera body.

### 3.2 Summary of the Analysis

Table II recaps the failure modes and the involved camera components. Further, we add a discussion on possible mitigations that can be implemented in the camera for each failure.

TABLE II
FAILURE MODES, FAILURES APPORTIONMENT TO THE CAMERA COMPONENTS, AND DISCUSSION ON MITIGATIONS.

| Failure | Comp. | Identified mitigations and practices |
|---|---|---|
| BRIGHT | Lens | Brightness, if detected, can be corrected to a certain extent at post-processing. Entirely black or entirely white images can be detected easily, but it is obviously difficult to recover the original image. |
| BLUR | | Various methods of removing or correcting image blurring exist [45], [46]. For example, Cai et al. [45] formulates the blind blurring as a new joint optimization problem, which simultaneously maximizes the sparsity of the blur kernel and the sparsity of the clear image. |
| BRLE | | A broken lens may be detected through image processing, but it may be difficult to re-create a clean image. |
| BRVR | | Methods similar to blur removal [45], [46] are applicable for this failure. |
| DIRTY | | Image processing solutions can remove localized rain and dirt effects from a single image [48]. For example, physics-based methods [47] can remove dust and dirt effects from digital photographs and videos. |
| FLARE | | Lens artifacts like lens flare and ghosting can be reduced or removed at post-processing [49] from a single input image to restore the correct image. |
| RAIN | | For the mitigation of this failure, we refer to the same observations of Dirty and to [48]. |
| COND | Lens and Camera Body | Various works e.g., [50], [51], [52], address the problem of avoiding or removing condensation inside cameras. |
| HEAT | | Against desert and very cold temperatures, there are commercial solutions such as the video surveillance camera AXIS Q60-C PTZ [28] that meets the military standard MIL-STD-810G [60]. |
| SAND | | This failure needs to be prevented with proper casing of the camera. |
| ICE | | Camera devices to heat lenses [53] prevent or at least reduce condensation of moisture. |
| WATER | | This failure needs to be prevented with proper casing of the camera. |
| BRACK | | Approaches exist specifically to prevent corrosion of surfaces in contact with seawater, brackish water, or fresh water [54]. |
| WIND | | |
| ELOV | Cam. Body | These failures need to be prevented e.g., with proper casing of the camera, reliable control circuits and reliable sensors. |
| NBAYF | Bayer Filter | |
| BAND | Image Sensor | There are various ways to reduce visual effects of banding e.g., by applying dithering patterns [59]. |
| BRLINES | | This failure is generally mitigated by the current knowledge and technology, and it is now rare. |
| DEAPIX | | It is possible to detect dead pixels with solutions implementable directly in the embedded device [57]; circuits to correct dead pixels can be fabricated on a single integrated chip [58]. |
| SPOTS | | Many image processing methods remove blemishes e.g., Zamfir et al. [55] detect blemishes and compute their physical appearance (size, shape, position and transparency) in an image as a function of camera settings, while Steinberg et al. [56] process images to automatically correct dust. |
| NOACT | ISP | It may be easy to detect this failure at system level, but it is not possible to recover the image. |
| CHRO-MAB | | Chromatic aberration effects can be reduced e.g., with image processing [22], [67]. |
| DEMOS | | Efficient ways for demosaicing exist as [61], [62]; however, it is difficult to recover the image in case of demosaicing failure. |
| DISTOR | | Lens distortion can be measured and detected [65], and it can be corrected with image processing [64], [66]. |
| NOISE SHARP | | Several solutions to reduce or remove noise and sharpness are available, also in commercial tools. Solutions that operate at sensor-level also exist e.g., [63], [29]. |

## 4 EFFECTS ON DETECTION APPLICATIONS

We study the effects of failures by exercising six object detectors on images captured by an RGB vehicle camera. The images are from the KITTI Vision Benchmark Suite [75], [76] for autonomous driving. The object detectors are selected amongst solutions for mono and stereo cameras that are enlisted in the rankings available on the KITTI website [92]. To perform the injections, we developed a Python library available at [9]. All the failed images and the detection results are available at [10].

### 4.1 KITTI image dataset and target metrics

The KITTI Vision Benchmark Suite (hereafter, just KITTI) includes images showing a variety of street situations captured from a moving platform driving around the city of Karlsruhe. A set of 7481 labeled images is provided with the ground truth, defined through labels that contain the bounding boxes of objects like cars, pedestrians, cyclists amongst others. As all images are captured by a stereo camera (i.e., two parallel sets of images are provided), they can be used for both 2D and 3D object detection. A similar set of images is intended for fair testing and comparison of object detectors, and ground-truth labels are not released.

KITTI includes a benchmark to measure object detection performance based on the average precision AP [77]. AP summarizes the shape of the precision/recall curve, and it is defined as the mean precision at a set of $n$ equally spaced recall levels. Value $n$ is set to 11 in [77], while it is set to $n=40$ in KITTI. In other words, we have that:

$$AP = \frac{1}{40} \sum_{r \in \{0, \frac{1}{40}, \frac{2}{40}, \cdots, 1\}} p_{interp}(r)$$

where the precision $p$ at each recall level $r$ is interpolated by taking the maximum precision measured for which the corresponding recall exceeds $r$ [77]. We compute AP using the standard KITTI configuration for cars detection, which confirms detection if there is at least an overlap of 70% between the ground-truth bounding boxes and the generated bounding boxes.

### 4.2 The selected object detectors

We selected six object detectors by exploring the ranking for 2D and 3D object detection available on the KITTI website. The mandatory criteria for the selection are: i) source code or trained model publicly released; ii) usage of RGB cameras only (no point clouds); iii) not older than five years i.e., 2016; iv) output must include bounding boxes in the KITTI format. With this last criterion, we can sample the individual image outputs to verify that low APs are caused by wrong detections and not because of processing issues.

We select four object detectors for mono cameras (FastBox, SqueezeDet, SqueezeDet+, ResNet50), and two for stereo cameras (DSGN, Disp R-CNN). In the following description of the object detectors, we report on the image data augmentation used during training. In fact, we may conjecture that some augmentation techniques may increase robustness against specific camera failures, because they add effects analogous to some of the failures (intuitively, in an analogous way as in adversarial training to defend against adversarial images [95]).

FastBox [81] is designed to achieve a high detection performance at a very fast inference speed, and it is able to jointly perform road segmentation, car detection and street classification. We apply the KittiBox scripts at [82] to train and execute FastBox. Regarding image data augmentation, color features are augmented by applying random brightness and random contrast to the images, while spatial features are distorted by applying random flip, random resize, and random crop.

SqueezeDet and SqueezeDet+ [83] are designed to be small and fast, with also a focus on energy efficiency. They are both convolutional neural networks based on SqueezeNet [84]; SqueezeDet+ is a variation of SqueezeDet, where a more powerful SqueezeNet is adopted. ResNet50 is a variant of the above, where the ResNet50 [85] network is used. Image data augmentation techniques used on SqueezeDet, SqueezeDet+ and ResNet50 are only random cropping and flipping.

DSGN [87] is an end-to-end stereo-based 3D object detection pipeline, which relies on space transformation from 2D features to achieve a 3D structure. The image data augmentation used is horizontal flipping only.

Disp R-CNN [88], which also operates on stereo cameras, considers 3D object localization as a learning-aided geometry problem rather than an end-to-end regression problem. It extends stereo-mask R-CNN [90], and its main characteristic is that it treats the object Region of Interest [89] as an entirety rather than independent pixels. The only reported image data augmentation is the image flip of stereo-mask R-CNN.

### 4.3 Failures configuration

From the failure modes in Sect. 3, we define 130 failures configurations, reported in Table III. It should be noted that some failures similarly alter the image, and consequently we group failures that have similar effects on the output image. This applies to the Blurred and the Broken VR failures (we will consider only the first one, from now on), and to the Dirty and Spots failures (again, we will consider only the first one). Further, we ignore failures whose effect is either i) not providing an output image (Electrical Overload, No Action, Water), or ii) not univocally determined (Brackish/Salt-Water, Heat, Sand). We exclude Bright Lines and Wind as we believe they are very rare for vehicles cameras. We exclude No Lens Distortion correction as the images are not collected with wide-angle lenses.

### 4.4 Definition of the experimental campaign

First, we discuss the train/validation split of the KITTI image set, the training of the object detectors, and the identification of the images on which we inject failures. The test set is not defined, because a separate test set is available for KITTI, whose labels are not disclosed: this is the same approach adopted in the papers describing the object detectors [81], [87], [83].

Concerning the four object detectors for mono cameras, to make the comparison as fair as possible we identify a set

TABLE III

FAILURES IMPLEMENTED, IMPLEMENTATION DETAILS, AND SELECTED CONFIGURATIONS.

| Failures | Implementation details | Configurations (with acronyms) |
|---|---|---|
| BAND | We use the PIL Image Module [12] resize() and blend() methods to superimpose images with dark stripes in [9] | BAND1, BAND2: two different banding effects are applied, by superimposing two different images with dark stripes |
| BLUR (and BRVR) | Blur is added using cv2 [11] method $cv2.blur(img,(i,i))$, where $i$ is the blur level | BLUR: we try 25 different levels of blur, with $(i,i)$ ranging from (1,1) i.e., very light blur, to (25,25) that is a very strong blur |
| BRIGHT | Brightness is modified using PIL $ImageEnhance$ module ($ImageEnhance.Brightness$) [12] | BRIGHT: we experiment with 10 different levels of brightness ranging from 0 (black image) to 15 (almost white): [0, 0.3, 0.6, 1.5, 3, 4.5, 6, 7.5, 10, 15] (note that the clean image is when the brightness level is 1). |
| BRLE | Failures are simulated using the PIL Image Module [12] and images of broken lens, dirt, rain drops, condensation, ice on lens that are superimposed to the original image. This is done either using PIL $paste$ or $blend$ methods, respectively if the image to be superimposed is transparent or not. PIL $resize$ and brightness adjustments are used as needed, to achieve the required effect | BRLE: we try 16 configurations, by superimposing 16 different images of scratched lens from [91] |
| COND | | COND: we try 3 configurations, by superimposing 3 transparent images representing condensation on lens available at [9] |
| DIRTY (and SPOTS) | | DIRTY: we try 36 configurations, by superimposing 36 different images representing dirt on lens from [91] |
| ICE | | ICE: we try 4 configurations, by superimposing 4 different transparent images representing ice on lens available at [9] |
| RAIN | | RAIN: we try 5 configurations, by superimposing 5 different transparent images representing rain on lens available at [9] |
| CHROMAB | Image processing with PIL Image Module [12], numPy [13] and [68]. | CHROMAB1-b, CHROMAB1-nb, CHROMAB2-b, CHROMAB2-nb: we introduce the effect of Chromatic Aberration for two incremental levels of aberration, with and without blur |
| DEAPIX | We change the color channel with cv2 [11]. It is sufficient to set to (0,0,0) the image pixels that should appear black | DEAPIX1, DEAPIX50, DEAPIX200, DEAPIX500: we consider respectively a single black pixel, 50, 200 and 500 black pixels randomly scattered on the image<br>DEAPIX-vcl: one black vertical line is drawn at the center of the image<br>DEAPIX-3l: 2 horizontal black lines and 1 vertical black line are drawn |
| DEMOS | We create an image twice the size of the original one, and we map the RGB values from the original image according to the BGGR pattern in Fig. 2 | DEMOS: the image is created twice the $(h, w)$ dimensions of the original image, with colors decomposed in a BGGR pattern |
| FLARE | We use flare effects from the Automold [86] library | FLARE: we add flare with random position and angle |
| NBAYF | We convert the image to a numPy array and then we multiply it by [0.2989, 0.5870, 0.1140], which are weights for RGB-to-greyscale conversion [79] | NBAYF: We multiply all RGB pixels of the selected images by the scaling factor [0.2989, 0.5870, 0.1140] |
| NOISE | Speckle noise is introduced using cv2 [11] and numPy's np.random.normal(0, $i$, img.size) [13], with $i$ representing the noise level | NOISE: we add gaussian noise on the entire image, applying 10 different values for the Standard deviation (spread or "width") of the distribution, with $i$ ranging from 0.2 (very low noise) to 5 (excessive disturbance) |
| SHARP | Sharpness is removed using the PIL $ImageEnhance.Sharpness$ [12] | SHARP: the method is invoked with 6 different sharpness factors, ranging from -5 (severe loss of sharpness) until 0 (visual effect is similar to a light blur) |

of images that is not used in the train split. We observe that Fastbox is trained in [81] with a train/validation split of [7001, 480] images, while the remaining three object detectors are trained in [83] with a train/validation split of [3741, 3740] images. Consequently, we extract 480 images from the 7481 labeled KITTI images: these images are the validation split of FastBox, and they are also included in the 3740 images of the validation split of the remaining three object detectors. Then, all object detectors are retrained. We stop the training when we reach scores analogous to those presented by their authors in the respective papers. We will inject failures on the 480 images previously selected. Object detectors for stereo cameras are instead not retrained because of computational load issues. In this case, we use the trained models provided by the authors (which relied on a train/validation split of [3712, 3769] images following [94]), and we select 500 images on which we will inject failures.

To explain the experimental campaign, let us consider a single object detector. Each of the 130 failure configurations of Table III is applied on the images, selected as described above. This leads to 130 sets of altered images, which are stored in 130 folders, plus one folder with the clean images (for a total of 131 folders). Note that for stereo cameras, we simulate the failure of one of the two cameras i.e., we inject failures in only one of the two parallel sets of images. Then, the object detector is applied on such 131 folders i.e., it performs the detection on the images in each folder and saves outputs in temporary folders. It should be noted that for each processed image, an object detector outputs a text file containing the identified bounding boxes; this leads to 131 folders each containing various text files. To compute the AP, these text files need to be further processed: we either rely on utilities of the object detector when available (this is the case for squeezeDet, squeezeDet+, resnet50), or we use the tool at [80] (this is the case for KittiBox, DSGN and Disp R-CNN).

TABLE IV
RESULTS ON CLEAN IMAGES AND MAXIMUM AP.

| Algorithm | Metric | AP on clean images (%) | Max AP (%) | Failure configuration for Max AP |
|---|---|---|---|---|
| FastBox | 2D AP | 86.63 | 87.25 | DIRTY |
| SqueezeDet | 2D AP | 86.49 | 87.05 | BAND1 |
| SqueezeDet+ | 2D AP | 88.82 | 89.15 | DIRTY |
| ResNet50 | 2D AP | 87.90 | 88.33 | DIRTY |
| DSGN | 2D AP | 89.67 | 91.81 | BLUR (2,2) |
| DSGN | 3D AP | 70.80 | 71.18 | BRLE |
| Disp R-CNN | 2D AP | 89.94 | 90.08 | SHARP 0 |
| Disp R-CNN | 3D AP | **57.89** | **63.14** | **DEAPIX-vcl** |

TABLE V
AP (IN PERCENTAGE) FOR BROKEN LENS, DIRTY AND RAIN.

| | AP | BRLE % min | BRLE % max | BRLE % avg | DIRTY % min | DIRTY % Max | DIRTY % avg | RAIN % min | RAIN % max | RAIN % avg |
|---|---|---|---|---|---|---|---|---|---|---|
| FastBox | 2D | 68 | 87 | 83 | 78 | 87 | 86 | 38 | 87 | 71 |
| SqueezeDet | 2D | 69 | 86 | 83 | 74 | 87 | 84 | 33 | 86 | 67 |
| SqueezeDet+ | 2D | 72 | 89 | 87 | 78 | 89 | 88 | 41 | 88 | 74 |
| ResNet50 | 2D | 60 | 87 | 82 | 78 | 88 | 86 | 21 | 87 | 67 |
| DSGN | 2D | 87 | 90 | 89 | 88 | 90 | 90 | 81 | 90 | 87 |
| DSGN | 3D | 64 | 71 | 69 | 66 | 71 | 69 | 50 | 70 | 62 |
| Disp R-CNN | 2D | 81 | 90 | 89 | 89 | 90 | 90 | 71 | 90 | 84 |
| Disp R-CNN | 3D | 48 | 58 | 56 | 53 | 63 | 57 | 27 | 58 | 48 |

To guarantee reproducibility of results, all the altered images are available at [10], and the software to apply failures configuration is at [9]. All computations are performed on a Dell Precision 5820 Tower with a 12-Core I9-9920X and GPU Nvidia Quadro RTX5000.

### 4.5 Analysis of results

We discuss the impact of failures on the object detection task. We remark that our objective is to show that camera failures have detrimental effects on different object detectors (and consequently should always be considered a credible threat), while comparing object detectors is not the scope of this work.

We first review results on clean images, which are reported in Table IV and constitute the baseline of our analysis. Object detectors DSGN and Disp R-CNN compute both bi-dimensional (2D) and tri-dimensional (3D) bounding boxes, measuring respectively 2D AP and 3D AP. Interestingly, Table IV shows that AP on clean images is not the highest AP, with the exception of SqueezeDet. This may raise the question of whether object detectors benefit from certain failure configurations. We can disregard any systematic benefit, because the improvements in Table IV are just small fluctuations on the AP score, with the (partial) exception of DEAPIX-vcl on Disp R-CNN under the DEAPIX-vcl failure, which raises 3D AP from ~58% to ~63%.

We now discuss the failures from Table III. We organize the discussion based on the apportionment of failures to the camera and its components, following Table I and Table

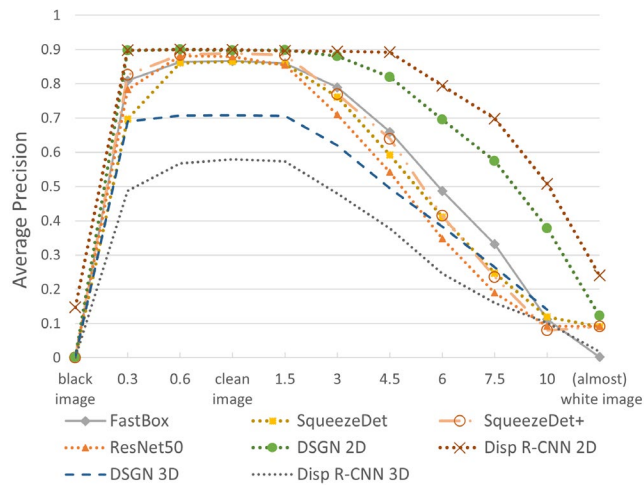

Fig. 5. AP measured for 10 different brightness alterations.

II (clearly, we ignore the cases in which there is no output image, for example the ice failure in the camera body). Consequently, we discuss: i) lens failures; ii) failures that can happen both in the lens and the camera body; iii) Bayer filter failure; iv) image sensor failures; v) ISP failures. Last, we comment on the possible benefits of data augmentation.

**Analysis of lens failures**. We discuss the simulated failures of Table III that affect lenses only i.e., brightness, blurred, broken lens, dirty, flare, rain.

Fig. 5 shows the AP under 10 different brightness failures. As expected, significant alterations of brightness drastically reduce the detection capability. This behavior is also common to DSGN and Disp R-CNN: the object detectors require that both input images are well-elaborated, and their detection performance is severely affected by the failure of just one camera. We obtain a similar effect with the 25 configurations of blurred failure (plot is not shown for brevity). For example, we can consider BLUR(10, 10), which is the blur level of Fig. 4c. No object detector has AP above 0.6, even if the vehicles are visually distinguishable.

We comment instead on broken lens, dirty and rain failures with the help of Table V. If the AP deviates of less than 5% from the AP on clean images, the corresponding cell in Table V is colored. We can deduce that most of the BRLE (scratches) and DIRTY failures are well-tolerated by most of the object detectors. These failures are simulated by overlaying images: clearly, the characteristics of the image that is overlayed have a relevant role on the measured AP e.g., because of the amount and size of raindrops contained.

Last, flare effects reduce AP to varying degrees. AP of object detectors for the mono camera is significantly reduced, with FastBox being the least affected (AP is reduced from 87% to 60%), and SqueezeDet+ the most affected (AP drops from 89% to 57%). DSGN and Disp R-CNN appear overall resilient to flare effects (the AP is reduced by at most 13 points; this happens with Disp R-CNN 3D which drops from 58% to 45%). This is probably because they rely on two cameras and this can partly mitigate our injected flare effect, which in many cases affects only a small part of the image.

**Analysis of lens-camera body failures.** The simulated failures that affect both lens and camera body are condensation and ice failures. The reduction of AP depends strongly on the overlayed figure: if the ice crystals and the condensation cover a significant part of the image, the

 

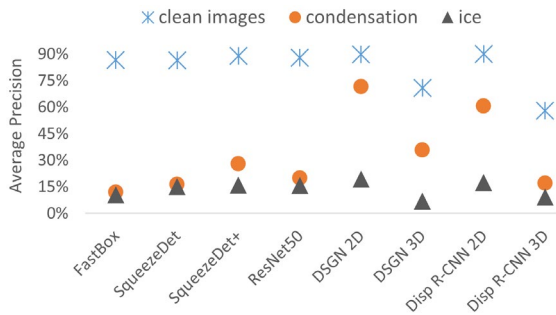

Fig. 6. Minimum AP with ice and condensation failures.

ability to detect objects is altered. In Fig. 6 we show the minimum AP recorded under ice and condensation failures. All object detectors are severely affected by these failures. DSGN and Disp R-CNN can tolerate the condensation failures more than ice failures, most likely because the condensation has higher transparency.

**Analysis of Bayer filter failures.** We consider the No Bayer Filter failure. This showed a peculiar behavior: the removal of colors according to a correction factor has a non-negligible impact on the various detectors, leading to a massive reduction of AP, which is measured below 10%. Additionally, the execution of Disp R-CNN failed under such configuration. The exception is DSGN which just reduces 2D AP by 3% and 3D AP by 19%. We observe that none of the object detectors was trained with color augmentation.

**Analysis of image sensor failures.** Following Table III and Table II, the simulated failures that affect both lens and camera body are banding and dead pixels failures.

The two configurations of the banding failure BAND1 and BAND2 are well-tolerated by the object detectors, leading to minimal AP reduction. The worst case is with ResNet50, where the AP is reduced from 88% with clean images to 84% in the case of BAND2. Also, dead pixel failures have limited effects on the image processing phase. In fact, for all the configurations DEAPIX1, DEAPIX50, DEAPIX200, DEAPIX500, DEAPIX-vcl, DEAPIX-3l, the worst AP reduction is with ResNet50, that is also in this case from 88% on clean images to 84% with the DEAPIX500 failure.

**Analysis of ISP failures.** Failures of the ISP that we simulate are no chromatic aberration, no demosaicing, no noise reduction, no sharpness correction.

The failure no chromatic aberration correction is a serious concern as it generally leads to a severe loss in AP, also for stereo cameras. Table VI marks in dark gray the cases in which AP is reduced by above 50%, and in light grey the AP reductions between 20% to 50%.

To simulate the no demosaicing failure, we revert the action of the ISP when producing the jpeg from the input raw. We achieve this by decomposing each pixel in its individual RGB contribution, as described in Table III. In this way, an image of larger scale than the clean image is created. This is the reason some object detectors are not able to process the images produced when injecting this failure. Disp R-CNN rates 89,46% 2D AP and 43,30% 3D AP, showing limited AP reduction with respect to the clean images; the other object detectors score 0% AP, except for DSGN which fails during execution, for the reasons explained above.

Last, the two failures no noise reduction and no sharpness correction show predictable trends: incremental values of NOISE and SHARP slowly degrade performances. NOISE progressively reduces AP as can be seen in Fig. 7. Instead, some object detectors can tolerate unsharpened images: even with SHARP set to -5, FastBox, SqueezeDet, DSGN 2D and Disp R-CNN 2D still show AP ≥ 70%.

**On data augmentation.** An additional consideration is provided about the possible use of image data augmentation strategies, to understand if it may improve robustness. As reported above, FastBox has been trained with random brightness, contrast, flip, resize and crop. Instead, ResNet50, SqueezeDet, SqueezeDet+ apply random cropping and flipping, while DISP-RCNN and DSGN apply image flipping. FastBox is slightly more robust than ResNet50, SqueezeDet, SqueezeDet+ under the brightness failure. The other data augmentation strategies are instead not relatable to the injected failures. While this is a hint to build our future works, this analysis is insufficient to conclude on the usefulness of data augmentation to improve robustness against camera failure, and more specific tests are needed.

**Summary of the analysis**. None of the object detectors is significantly more robust to camera failures than the others. Even if the object detectors are different, they are all affected by the simulated failures to a relevant extent. This underlines the need to carefully consider camera failures and their possible effect when deploying object detectors.

TABLE VI
AP (IN PERCENTAGE) FOR NO CHROMATIC ABERRATION (VALUES IN PERCENTAGE).

| AP (%) | | CHROM AB1-nb | CHROM AB1-b | CHROM AB2-nb | CHROM AB2-b |
|---|---|---|---|---|---|
| FastBox | 2D | 23.03 | 9.58 | 6.98 | 0.81 |
| SqueezeDet | 2D | 42.16 | 20.72 | 23.86 | 5.01 |
| SqueezeDet+ | 2D | 52.79 | 27.66 | 29.89 | 13.53 |
| ResNet50 | 2D | 56.44 | 18.35 | 34.08 | 6.94 |
| DSGN | 2D | 58.43 | 58.97 | 23.56 | 29.44 |
| | 3D | 1.17 | 1.00 | 0.33 | 0.68 |
| Disp R-CNN | 2D | 79.76 | 60.51 | 67.89 | 37.63 |
| | 3D | 2.45 | 3.03 | 9.09 | 9.09 |

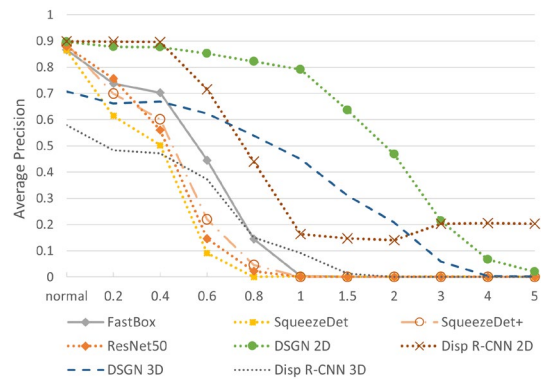

Fig. 7. AP measured for 10 different noise alterations.

## 5 EFFECTS ON DRIVING TASKS

To underline the possible impact on safety, we inject the failures in the vehicle camera of an autonomous driving simulator. Logs and some videos that describe the various runs are available at [10].

### 5.1 Autonomous Driving with Carla simulator

We opt for the Open Urban Driving Simulator Carla (Car Learning to Act, [1]) to create a vehicle that is autonomously driving in a town using only the RGB camera.

Carla has been implemented as an open-source layer over the Unreal Engine 4 (UE4, [2]) to support training, prototyping, and validation of autonomous driving models, including both perception and control. Carla includes urban layouts, several vehicle models, buildings, pedestrians, street signs, etc. Further, it provides information on the simulated vehicles as position, orientation, speed, acceleration, collisions, and traffic violations.

Amongst the autonomous driving agents that exist for Carla, we use the trained agent from [3]. Technical details on the trained agent are outside the scope of this paper and are reported in [3]; we introduce only the notions required to illustrate our tests. Using this trained agent, at each simulation step it is acquired: i) one RGB image from the frontal camera of the vehicle at a resolution of 384×160 pixels (this is significantly smaller than the KITTI images used in Section 4), and ii) the current speed from the speed sensor. These values are processed by the trained agent to predict waypoints in the camera coordinates, and then these waypoints are projected into the vehicle's coordinate image [3]. In simpler words, the trained agent "designs" a trajectory composed of five waypoints on the image acquired from the camera. From this, a low-level controller is executed that decides the steering angle, the throttle level, and the braking force. Finally, throttle, speed, and braking are applied to the vehicle.

We selected this trained agent amongst the various available because: i) it uses only the camera as sensing system; ii) it presents very good performances, with a minimal number of collisions. Further, the model was trained with image data augmentations following [8], including pixel dropout, blurring, Gaussian noise, and color perturbations which partially overlap with our set of failures.

### 5.2 Injection strategy and failure implementation

Our injection strategy consists of the following actions, performed at each simulation step: i) acquire the output image from the camera; ii) modify the image by injecting the selected failure before the trained agent processes the image; and iii) feed the modified image to the trained agent. We report on the following configurations from Table III: BAND1, BRIGHT 0 (black images), BRIGHT 1.5, BLUR (12,12), DEAPIX1, DEAPIX200, DEAPIX-vcl, NBAYF, CHROMAB2-nb, NODEMOS, NOISE 1, SHARP - 3.5, one BRLE, one COND, one DIRTY, one ICE, one RAIN. These configurations are selected to describe the behavior of the trained agent under different camera failures. The images overlayed for these tests are available at [9]. We exclude Flare as it depends on the sun position and it needs to consider vehicle movements, requiring environmental data; further, it would overlap with the flare effects already represented in the Carla simulator.

### 5.3 Test plan and execution

The test plan is based on the *corl2017* benchmark from [1]. The test plan is composed of multiple runs in which a target vehicle must reach a destination position $B$ from a starting position $A$ before a timeout expires. The timeout value is the time required to cover the distance from $A$ to $B$ at an average speed of 10 Km/h as in [1], [3]; this matches the settings of the corl2017 benchmark, which takes into account vehicle stops at traffic lights and traffic. Re-using the nomenclature from [1], the starting position $A$ and the destination position $B$ are selected such that three *test objectives* are set:

- *Straight road*: Destination position $B$ is located straight ahead of the starting position $A$.
- *Turn road*: Destination position $B$ is one turn away from the starting position $A$.
- *Navigation*: There is no restriction on the location of the destination position $B$ relative to the starting position $A$; this results in runs of longer distance and multiple turns.

For each run, the success criterion is to reach destination $B$ before the expiration of the timeout. The failure criterion is whenever the vehicle collides or the timeout expires: we modified the *corl2017* benchmark to halt the run whenever a collision occurs, as in our work we prioritize safety over traveled distance.

The target town we select is Carla Town02, which is a basic town layout with all "T junctions" and it is also the town used for testing the trained agent in [3]. Further, we select three different weather conditions: clear noon, wet cloudy noon, and hard rain sunset. In addition, in each of the runs performed, the town includes exactly 50 vehicles and 30 pedestrians. We always use the same randomization seed so that the spawning positions of vehicles and pedestrians are the same in the repeated runs.

The experiments were organized in two phases. In the first phase, we performed golden runs on clean images, i.e., we execute the simulation runs without introducing any modification to the images captured by the camera. The second phase repeats the same runs of the previous phase, but with the injection of camera failures to each acquired image.

The three test objectives are investigated in 150 runs (50 runs for each test objective) for the golden runs and for the 17 failure configurations that we inject. With a total of 2700 runs, the simulated time in Carla corresponds to approximately 80 hours of driving. The simulations were executed on a Dell Precision 5820 Tower with a 24-Core I9-9920X and GPU Nvidia Quadro RTX5000.

### 5.4 Collisions and success rate

First, we discuss the impact of each failure on the decisions of the trained agent of [3]. Fig. 8a and Fig. 8b show respectively the success rate and the number of collisions for the three test objectives Straight road, Turn road, and Navigation. Not surprisingly, runs on clean images perform the best, with the highest success rate and the lowest number

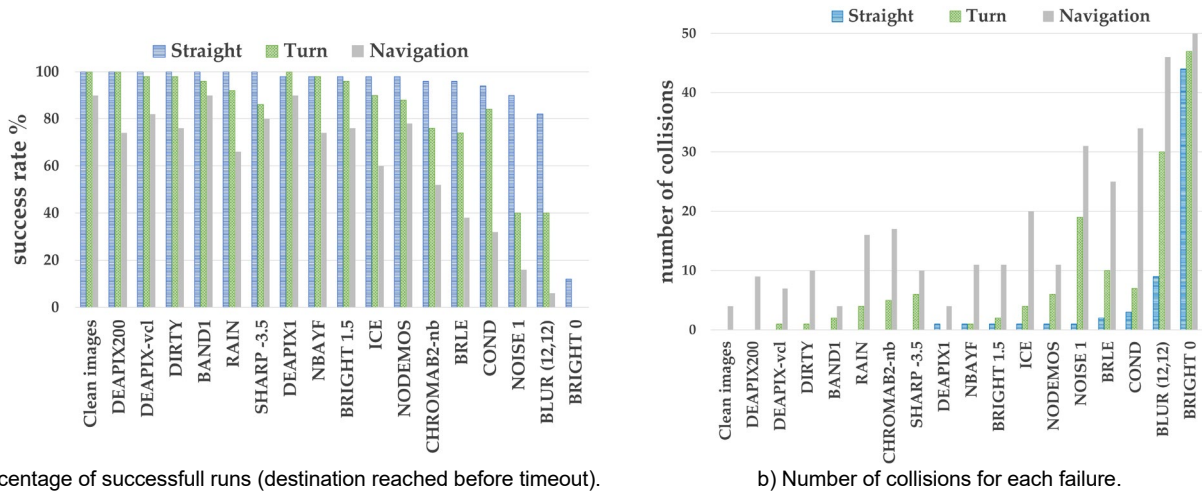

a) Percentage of successfull runs (destination reached before timeout).  
b) Number of collisions for each failure.

Fig. 8. Execution of the simulated runs in the three test objectives Straight road, Turn road, and Navigation road.

of collisions (4 collisions, all under the Navigation test objective). Similar results are achieved by DEAPIX1 (5 collisions) and BAND (6 collisions): differences from the golden runs are small.

As expected DEAPIX1 does not significantly affect the trained agent: this failure consists of a single black pixel introduced at the bottom right of the image. The trained agent is robust also against other dead pixel failures (DEAPIX200 and DEAPIX-vcl), and banding. These failures were also well-tolerated by the object detectors analyzed in Section 4. While this may suggest that these effects do not significantly alter the proper behavior of trained agents, it should be remarked that the results presented here cannot be generalized to any trained agent. For example, it has been proven that few altered pixels in strategic position can fail the classification task [96].

Failures BRIGHT 0, BLUR, NOISE 1, COND, BRLE are the worst-performing: these failures significantly modify the image. Still, they show some successful runs, and this may look especially surprising for the Straight road with BRIGHT 0 (image is black). This is simply because the car is moving forward blindly in a straight direction, and, if there are no obstacles, the run ends successfully.

The remaining failures have variable performances; however, for each of them, the number of collisions is always the highest in Navigation and the lowest in Straight road. The majority of collisions happen when the vehicle performs a turn, while instead the trained agent is generally able to avoid the vehicle in front. The image is sufficiently disrupted to not allow a proper calculation of the trajectory, but it still allows detecting objects right in front of the vehicle in a straight road.

Last, we observe that the order of failures in Fig. 8a and Fig. 8b is not the same. This is mostly because some runs terminate due to timeout: in general, a timeout occurs when the vehicle is unable to decide how to advance after a particular event. In fact, in few cases the vehicle stops to avoid a vehicle in a colliding trajectory: the trained agent can detect the object, so there is no collision but the driving does not restart.

**Impact on safety and generalization of results.** It is evident that our results depend on the target application, and consequently a univocal definition of failure criticalities and risks cannot be devised. However, these experiments bring evidence that the represented failures should not be ignored when building image-based autonomous driving systems and applications. We show that even failures with small visual effects affect the decisions of the trained agent, e.g., see the number of collisions under the DEAPIX200 failure that scatters 200 dead pixels on a camera with a resolution of 384×160 (above 60.000 pixels).

## 6. RELATED WORKS

To the best of our knowledge, no research works discuss the failure model of a vehicle camera, including an analysis of effects and safety risks, and the provision of a software library. However, several works deal with similar or inherent problems.

The *performance, robustness, and security of an RGB camera* have been widely explored, however usually focusing on specific elements or target metrics and without addressing the full set of failures. For example, Bijelic et al. [25] present a test and evaluation methodology to compare sensor technologies: the paper shows the difference between an image captured by a standard CMOS camera and one captured by a gated camera. Schops et al. [26], motivated by the limitations of existing multi-view stereo camera benchmarks (a stereo camera has two or more lenses, each with a separate image sensor: this allows the device to simulate binocular human vision and capture three-dimensional images [34]), introduce a new dataset and a technique to minimizes the photometric errors. In [32] a simulation environment is presented which includes the virtual structures of a car designed for autonomous driving tests; typical driving situations have been used to analyze how sensors respond when used in real circumstances as well as to confirm the impacts of environmental conditions. Considering instead sensor security issues, Petit et al. [19] blind a commercial camera system used in commercial vehicles with several light sources. The work shows that leveraging a laser or LED matrix could blind the camera. Similarly, Yan et al. [34] successfully blind the camera by aiming the LED and the laser light at the camera directly: radiating a laser beam against a camera of a vehicle may cause irreversible damage and disrupt the corresponding autonomous applications. In general, it is observed that, because of the vulnerability of the camera caused by its optical characteristics, it is difficult to build a completely secure camera system [18].

In the domain of image-based AI/ML algorithms and applications, many works acknowledge that *the risk of accidental alterations of the output image of the camera is realistic* e.g., [47]. However, this consideration is usually ancillary to the main contribution of the work. Nonetheless, works in the AI/ML domain strongly helped us refine and cross-check the completeness of the failure modes we identified. In fact, chromatic aberration, noise, color temperature, blur and brightness alteration are often considered in image-based AI/ML trained agents, although for the scope of data augmentation during training [30]. For example, Toromanoff et al. [31] present a new convolutional neural networks (CNN) model, in which label augmentation based on translation and rotation allows generating data using only a short-range fisheye (wide angle) camera. Menze et al. [23] elaborate a new model and data set for 3D scene flow estimation, and explicitly take advantage of the background movement caused by the camera mounted on a vehicle. Behzadan et al. [24] show a new deep reinforcement learning framework; to develop robust sensors and algorithms, testing under certain meteorological conditions is deemed crucial for determining the impact of bad weather on sensors.

Several other works instead focused on *security and robustness of the trained agents that contribute to the autonomous driving system*, trying to understand the possible modification of camera images that could be produced by an attacker, or to define corner cases. Attackers may maliciously alter the images with transformations that are similar, in concept, to those that could happen with non-malicious failures: also, these works were useful when identifying the set of failures. Most relevant, K. Pei et al. [35] apply input space reduction techniques to transform the image, and can simulate a wide range of real-world distortions, noises, and deformations. W. Wu et al. [36] present a faults model for deep neural networks classifiers, which includes several corner cases based on the alteration of the input image, including amongst the possible causes brightness, camera alignment, and object movements. Finally, evasion attacks consist in modifying the input to a classifier such that it is misclassified, while keeping the modification as small as possible [37]. For example, to create such adversarial images, the correct images can be modified by overlaying carefully crafted noise [38], altering few selected pixels [39], or with rotation and translation [40]. A representative list of such evasion attacks and their implementations is available at [37].

## 7. CONCLUSIONS

It is commonly acknowledged that RGB cameras may fail, and that a failed camera may produce altered images. This is a relevant issue if such images are further processed by safety-critical image-based applications. Nevertheless, to our knowledge, a thorough enumeration of camera failure modes and means for their reproduction are still missing. Especially when cameras are used for safety-critical applications e.g., in the autonomous driving domain, the definition of failure modes would benefit software and system engineers to build resilient architecture or to assess application robustness.

This paper identifies the failure modes of a vehicle camera, describing their effects on the output image. We believe this can benefit software and system engineers that can rely on a reference model both when architecting the system and when assessing robustness of intelligent systems. The discussion is complemented with the identification of potential mitigations, and with a software library that can be used to reproduce the failures on image sets. Further, we reproduced such failures in image-based AI/ML applications for autonomous driving (six car detectors and one self-driving agent), to understand the impact of failures. First, we showed that camera failures have a relevant impact on different object detectors, and consequently should always be considered a credible threat. Further, we showed that even failures that slightly perturb the image may alter the decisions of a trained agent. As possible detection strategies, we have hypothesized that data augmentation techniques may be a viable means to tolerate failures; while this is outside the scope of this paper, our current research works are in fact oriented in this direction. Last, we remark that our results are application-dependent: even if some failures are tolerated in our experiments, still we recommend to not disregard them when assessing robustness of image-based AI/ML applications.

## ACKNOWLEDGMENT

This work has been partially supported by the H2020 programme under the Marie Sklodowska-Curie grant agreement 823788 (ADVANCE) project and by the project POR-CREO SPACE "Smart PAssenger CEnter" funded by the Tuscany Region.

## REFERENCES

[1]   A. Dosovitskiy, G. Ros, F. Codevilla, A. Lopez, and V. Koltun, "CARLA: An Open Urban Driving Simulator," Conference on Robot Learning, pp. 1-16, 2017.

[2]   Unreal Engine, www.unrealengine.com [online].

[3]   D. Chen et al., "Learning by Cheating," Conference on Robot Learning (CoRL), pp. 66-75, 2019.

[4]   J. Huang et al., "Failure mode and effect analysis improvement: A systematic literature review and future research agenda," Reliability Engineering & System Safety:106885, 2020.

[5]   A. Bouti, and A. K. Daoud, "A state-of-the-art review of FMEA/FMECA," International Journal of reliability, quality and safety engineering 1.04:515-543, 1994.

[6]   E. Allen, and S. Triantaphillidou, "The manual of photography," CRC Press, 2012.

[7]   N. K. Guy, "The Lens: A Practical Guide for the Creative Photographer," Rocky Nook, Inc., 2012

[8]   F. Codevilla et al., "End-to-end driving via conditional imitation learning," IEEE International Conference on Robotics and Automation (ICRA). IEEE, 2018.

[9] Code to reproduce the image failures of this paper: https://github.com/francescosecci/Python_Image_Failures [online]

[10] Archive of altered images and collected results: https://drive.google.com/drive/folders/1M5itBeFA6p7-pY-jSuVZ3IFBi6fMNIfek?usp=sharing [online].

[11] CV2 filtering, https://docs.opencv.org/2.4/modules/imgproc/doc/filtering.html [online].

[12] PIL 3.0, https://pillow.readthedocs.io/en/3.0.x/ [online].

[13] NumPy, https://numpy.org/ [online].

[14] S.-H. Teng, and S.-Y. Ho, "Failure mode and effects analysis: an integrated approach for product design and process control," International journal of quality & reliability management, 13(5):8-26, 1996.

[15] C. Premebida, G. Melotti, and A. Asvadi, "RGB-D Object Classification for Autonomous Driving Perception," RGB-D Image Analysis and Processing. Springer, Cham, pp. 377-395, 2019.

[16] J. Levinson et al., "Towards fully autonomous driving: Systems and algorithms," 2011 IEEE Intelligent Vehicles Symposium (IV). pp. 163-168, IEEE, 2011.

[17] S.-C. Lin et al., "The architectural implications of autonomous driving: Constraints and acceleration," Proceedings of the 23rd International Conference on Architectural Support for Programming Languages and Operating Systems, pp. 751-766, 2018.

[18] K. Ren et al., "The Security of Autonomous Driving: Threats, Defenses, and Future Directions," Proceedings of the IEEE, pp. 357-372, 2019.

[19] J. Petit et al., "Remote attacks on automated vehicles sensors: Experiments on camera and lidar," Black Hat Europe 11, 2015.

[20] H. Hagiwara, "Zoom lens and electronic apparatus," U.S. Patent No. 9.541.768, 10 Jan. 2017.

[21] C. Hughes et al., "Wide-angle camera technology for automotive applications: a review," IET Intelligent Transport Systems 3.1: 19-31, 2009.

[22] S.-W. Chung, B.-K. Kim, and W.-J. Song, "Removing chromatic aberration by digital image processing," Optical Engineering 49.6:067002, 2010.

[23] M. Menze, and A. Geiger, "Object scene flow for autonomous vehicles," Proceedings of the IEEE conference on computer vision and pattern recognition, pp. 3061-3070, 2015.

[24] V. Behzadan, and A. Munir, "Adversarial reinforcement learning framework for benchmarking collision avoidance mechanisms in autonomous vehicles," IEEE Intelligent Transportation Systems Magazine, vol. 13, no. 2, pp. 236-241, 2019.

[25] M. Bijelic, T. Gruber, and W. Ritter, "Benchmarking image sensors under adverse weather conditions for autonomous driving," 2018 IEEE Intelligent Vehicles Symposium (IV), pp. 1773-1779, IEEE, 2018.

[26] T. Schops et al., "A multi-view stereo benchmark with high-resolution images and multi-camera videos," Proceedings of the IEEE Conference on Computer Vision and Pattern Recognition, pp. 3260-3269, 2017.

[27] J. B. Phillips, and H. Eliasson, "Camera image quality benchmarking," John Wiley & Sons, 2018.

[28] AXIS Q60 PTZ, https://www.axis.com/it-it/products/axis-q60-series [online].

[29] L. Guo et al., "Automatic sensor correction of autonomous vehicles by human-vehicle teaching-and-learning," IEEE Transactions on Vehicular Technology 67.9: 8085-8099, 2018.

[30] A. Carlson et al., "Modeling Camera Effects to Improve Visual Learning from Synthetic Data," Proceedings of the European Conference on Computer Vision (ECCV) Workshops, 2018.

[31] M. Toromanoff et al., "End to end vehicle lateral control using a single fisheye camera," IEEE/RSJ International Conference on Intelligent Robots and Systems (IROS), pp. 3613-3619, IEEE, 2018.

[32] S. Hossain et al., "CAIAS simulator: self-driving vehicle simulator for AI research," International Conference on Intelligent Computing & Optimization, pp. 187-195, Springer, Cham, 2018.

[33] Z. Zheng, X. He, and J. Weng, "Approaching camera-based real-world navigation using object recognition," Procedia Computer Science 53: 428-436, 2015.

[34] C. Yan, W. Xu, and J. Liu, "Can you trust autonomous vehicles: Contactless attacks against sensors of self-driving vehicle," DEF CON 24.8: 109, 2016.

[35] K. Pei et al., "Towards practical verification of machine learning: The case of computer vision systems," arXiv preprint arXiv:1712.01785, 2017.

[36] W. Wu et al., "Deep validation: Toward detecting real-world corner cases for deep neural networks," 49th Annual IEEE/IFIP International Conference on Dependable Systems and Networks (DSN), pp. 125-137, IEEE, 2019.

[37] M.-I. Nicolae et al., "Adversarial Robustness Toolbox v1. 0.0.," arXiv preprint arXiv:1807, 2018.

[38] S. M. Moosavi-Dezfooli, A. Fawzi, and P. Frossard, "Deepfool: a simple and accurate method to fool deep neural networks," Proceedings of the IEEE conference on computer vision and pattern recognition, pp. 2574-2582, 2016.

[39] D. V. Vargas, and S. Kotyan, "Robustness Assessment for Adversarial Machine Learning: Problems, Solutions and a Survey of Current Neural Networks and Defenses," arXiv preprint arXiv:1906.06026, 2019.

[40] L. Engstrom et al., "A rotation and a translation suffice: Fooling cnns with simple transformations," arXiv preprint arXiv:1712.02779 1.2:3, 2017.

[41] T. Northrup, "Tony Northrup's photography buying guide: How to choose a camera, lens, tripod, flash, & more," Waterford, Mason Press, 2016.

[42] P. Poli, "Fotografia digitale: guida completa," Apogeo, Milano, 2014.

[43] H. E. Townsend (ed.), "Outdoor atmospheric corrosion," ASTM, 2002.

[44] D. E. Elkins, "The camera assistant's manual," Taylor & Francis, 2013.

[45] J.-F. Cai et al., "Blind motion deblurring from a single image using sparse approximation," IEEE Conference on Computer Vision and Pattern Recognition, pp. 104-111, IEEE, 2009.

[46] L. Fang et al., "Separable kernel for image deblurring," Proceedings of the IEEE Conference on Computer Vision and Pattern Recognition, pp. 2885-2892, 2014.

[47] J. Gu et al., "Removing image artifacts due to dirty camera lenses and thin occluders," ACM Transactions on Graphics (TOG) 28.5: 1-10, 2009.

[48] D. Eigen, D. Krishnan, and R. Fergus, "Restoring an image taken through a window covered with dirt or rain," Proceedings of the IEEE international conference on computer vision, pp. 633-640, 2013.

[49] F. Chabert, "Automated lens flare removal," Technical report, Stanford University, Department of Electrical Engineering, 2015.

[50] M. Kondou, "Condensation prevention camera device," U.S. Patent No. 9.525.809, 20 Dec. 2016.

[51] D. Loiacono, "Noncondensing security camera housing window assembly," U.S. Patent Application No. 12/433.457, 2010.

[52] B. Englander, "Video camera unit, protective enclosure and power circuit for same, particularly for use in vehicles," U.S. Patent No. 5.455.625, 3 Oct. 1995.

[53] A. Talbert, B. M. Milford, and R. Nelson, "Lens heater," U.S. Patent No. 2.442.913, 8 Jun. 1948.

[54] W. J. Riffe, and J. D. Carter, "Method for the prevention of fouling and/or corrosion of structures in seawater, brackish water and/or fresh water," U.S. Patent No. 5.346.598, 13 Sep. 1994.

[55] A. Zamfir et al., "An optical model of the appearance of blemishes in digital photographs," Digital Photography III, Vol. 6502, International Society for Optics and Photonics, 2007.

[56] E. Steinberg, P. Bigioi, and A. Zamfir, "Detection and removal of blemishes in digital images utilizing original images of defocused scenes," U.S. Patent No. 7.295.233, 13 Nov. 2007.

[57] C.-Y. Cho et al., "Real-time photo sensor dead pixel detection for embedded devices," International Conference on Digital Image Computing: Techniques and Applications, pp. 164-169, 2011.

[58] K. Dong, "On-chip dead pixel correction in a CMOS imaging sensor," U.S. Patent No. 7.522.200, 21 Apr. 2009.

[59] S. Bhagavathy, J. Llach and J. f. Zhai, "Multi-Scale Probabilistic Dithering for Suppressing Banding Artifacts in Digital Images," IEEE International Conference on Image Processing, pp. 397-400, 2007.

[60] S. Barrett, "Environmental Engineering Considerations and Laboratory Tests," MIL-STD-810G, 2008.

[61] B. Grossmann and Y. C. Eldar, "An efficient method for demosaicing," 23rd IEEE Convention of Electrical and Electronics Engineers in Israel, pp. 436-439, Sep 2004.

[62] L. Shi, et al., "Demosaicing for RGBZ sensor," Computational Imaging XI, Vol. 8657, International Society for Optics and Photonics, 2013.

[63] Y.-M. Baek et al., "Noise reduction for image signal processor in digital cameras," 2008 International Conference on Convergence and Hybrid Information Technology, pp. 474-481, IEEE, 2008.

[64] S. Yoneyama et al., "Lens distortion correction for digital image correlation by measuring rigid body displacement," Optical engineering, 45.2:023602, 2006.

[65] B. Prescott, and G. F. McLean, "Line-based correction of radial lens distortion," Graphical Models and Image Processing 59.1: 39-47, 1997.

[66] H. S. Sawhney, and R. Kumar. "True multi-image alignment and its application to mosaicing and lens distortion correction," IEEE Transactions on Pattern Analysis and Machine Intelligence 21.3: 235-243, 1999.

[67] S. B. Kang, "Automatic removal of chromatic aberration from a single image," 2007 IEEE Conference on Computer Vision and Pattern Recognition, pp. 1-8, IEEE, 2007.

[68] Realistic Lens Blur/Chromatic Aberration Filter, https://github.com/yoonsikp/kromo [online]

[69] S. Lee, E. Eisemann, and H.-P. Seidel, "Real-time lens blur effects and focus control," ACM Trans. Graph. 29, 4, Article 65, 2010.

[70] F. F. Sabins, and J. M. Ellis, "Remote Sensing: Principles, Interpretation, and Applications," Fourth Edition, Waveland Press, 2020.

[71] D. Baumgartner et al., "Benchmarks of low-level vision algorithms for DSP, FPGA, and mobile PC processors," Embedded Computer Vision. Springer, London, 101-120, 2009.

[72] G. Sharma, "Photography Redefined," Lulu.com, 2013.

[73] S. C. Bhargava, "Electrical Measuring Instruments and Measurements," CRC Press, 2012.

[74] S. Kelby, "Scott Kelby's Digital Photography Boxed Set," Parts 1, 2, 3, 4, and 5, Peachpit Press, 2014.

[75] A. Geiger, P. Lenz, and R. Urtasun, "Are we ready for autonomous driving? The Kitti vision benchmark suite," 2012 IEEE Conference on Computer Vision and Pattern Recognition, pp. 3354-3361, IEEE, 2012.

[76] M. Menze, and A. Geiger, "Object scene flow for autonomous vehicles," Proceedings of the IEEE conference on computer vision and pattern recognition, pp. 3061-3070, 2015.

[77] M. Everingham, et al. "The pascal visual object classes (voc) challenge," International journal of computer vision 88.2:303-338, 2010.

[78] F. Secci, and A. Ceccarelli, "On failures of RGB cameras and their effects in autonomous driving applications," The 31st International Symposium on Software Reliability Engineering (ISSRE 2020), pp. 13-24, 2020.

[79] International Telecommunication Union (ITU), "BT.601: Studio encoding parameters of digital television for standard 4:3 and wide screen 16:9 aspect ratios," 2011.

[80] eval_kitti, https://github.com/cguindel/eval_kitti

[81] M. Teichmann et al., "Multinet: Real-time joint semantic reasoning for autonomous driving," IEEE Intelligent Vehicles Symposium (IV), pp. 1013-1020, IEEE, 2018.

[82] KittiBox, https://github.com/MarvinTeichmann/KittiBox [online].

[83] B. Wu et al., "Squeezedet: Unified, small, low power fully convolutional neural networks for real-time object detection for autonomous driving," Proceedings of the IEEE Conference on Computer Vision and Pattern Recognition Workshops, pp. 129-137, 2017.

[84] F. N. Iandola et al., "SqueezeNet: AlexNet-level accuracy with 50x fewer parameters and< 0.5 MB model size," arXiv preprint arXiv:1602.07360.

[85] K. He et al., "Deep residual learning for image recognition," Proceedings of the IEEE conference on computer vision and pattern recognition, pp. 770-778, 2016.

[86] Automold, https://github.com/UjjwalSaxena/Automold--Road-Augmentation-Library [online].

[87] Y. Chen et al., "Dsgn: Deep stereo geometry network for 3d object detection," Proceedings of the IEEE/CVF Conference on Computer Vision and Pattern Recognition, pp. 12536-12545, 2020.

[88] J. Sun et al., "Disp R-CNN: Stereo 3D Object Detection via Shape Prior Guided Instance Disparity Estimation," Proceedings of the IEEE/CVF Conference on Computer Vision and Pattern Recognition, pp. 10548-10557, 2020.

[89] K. He et al., "Mask r-cnn," Proceedings of the IEEE international conference on computer vision, pp. 2961-2969, 2017.

[90] P. Li, X. Chen, and S. Shen, "Stereo r-cnn based 3d object detection for autonomous driving," Proceedings of the IEEE Conference on Computer Vision and Pattern Recognition, pp. 7644-7652, 2019

[91] ActionVFX, https://www.actionvfx.com/ [online].

[92] KITTI web site, http://www.cvlibs.net/datasets/kitti/ [online].

[93] K. Hogan, "Options for Camera Raw in the Digital Workflow," SMPTE, pp. 1-18, 2014.

[94] X. Chen, et al. "3d object proposals for accurate object class detection," In Advances in Neural Information Processing Systems, pp. 424-432, 2015

[95] A. Madry, et al. "Towards deep learning models resistant to adversarial attacks." arXiv preprint arXiv:1706.06083, 2017.

[96] J. Su, D. Vargas, and K. Sakurai, "One pixel attack for fooling deep neural networks," IEEE Transactions on Evolutionary Computation 23.5 (2019): 828-841.

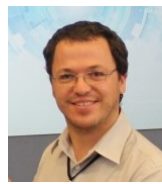

**Andrea Ceccarelli** is an Associate Professor in Computer Science at the University of Florence, Florence, Italy. His primary research interests are in the design, monitoring and experimental evaluation of dependable and secure systems, and systems-of-systems. He has been PC co-chair of the conferences SRDS and LADC, and he is a member of the IFIP WG 10.4 on "Dependable Computing and Fault-Tolerance". His scientific activities originated more than 100 papers which appeared in international conferences, workshops, and journals.

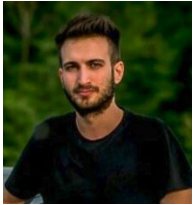 **Francesco Secci** graduated from the University of Florence, Florence, Italy, very recently, in 2020. After graduating, he moved to the industry where he is currently working in a book distribution company (Libro Co. Italia s.r.l.) with the qualification of software programmer for business management, for the data manipulation and the management of the company's visibility in the social field.